\documentclass{article}

\usepackage{graphicx}

\usepackage{hyperref}

\usepackage{amsmath}
\usepackage{amssymb}
\usepackage{mathtools}
\usepackage{amsthm}

\usepackage{natbib}
\usepackage{amsfonts,
        subcaption,
		url,
        float,
        markdown,
        comment,
        }
\usepackage{authblk}

\def \path {content/}

\theoremstyle{plain}

\theoremstyle{definition}

\theoremstyle{remark}

\title{Mixture of neural operator experts for learning boundary conditions and model selection}
\author[1]{Dwyer Deighan\thanks{dwyerdei@buffalo.edu}}
\author[2]{Jonas A. Actor\thanks{jaactor@sandia.gov}}
\author[2]{Ravi G. Patel\thanks{rgpatel@sandia.gov}}
\affil[1]{Department of Computer Science and Engineering, University at Buffalo, The State University of New York, Buffalo, NY, USA}
\affil[2]{Center for Computing Research, Sandia National Laboratories, Albuquerque, NM, USA}

\date{}

\begin{document}

\maketitle

\begin{abstract}
While Fourier-based neural operators are best suited to learning mappings between functions on periodic domains, several works have introduced techniques for incorporating non trivial boundary conditions. However, all previously introduced methods have restrictions that limit their applicability. In this work, we introduce an alternative approach to imposing boundary conditions inspired by volume penalization from numerical methods and Mixture of Experts (MoE) from machine learning. By introducing competing experts, the approach additionally allows for model selection. To demonstrate the method, we combine a spatially conditioned MoE with the Fourier based, Modal Operator Regression for Physics (MOR-Physics) neural operator and recover a nonlinear operator on a disk and quarter disk. Next, we extract a large eddy simulation (LES) model from direct numerical simulation of channel flow and show the domain decomposition provided by our approach. Finally, we train our LES model with Bayesian variational inference and obtain posterior predictive samples of flow far past the DNS simulation time horizon.\end{abstract}

\section{Introduction}\label{intro}

Fully resolved simulations of partial differential equations (PDEs) are prohibitively expensive for most systems, even on the largest supercomputers. Accordingly, under-resolved simulations are supplemented with models for subgrid-scale dynamics. For example, for the Navier-Stokes equations in the turbulent regime, the turbulence dynamics are approximated statistically via large eddy simulation (LES) or Reynolds averaged Navier-Stokes (RANS) models; such models are imperfect as they encode various empirical assumptions and hand-tuned approximations \cite{Pope_2000}.

Neural operators are a class of surrogate models that parameterize unknown operators with neural networks and can learn 
PDE models from high-fidelity simulation and/or experimental data. 
Originally introduced as a theoretical construction by \cite{Chen1995}, recent numerical implementations of neural operators have seen recent success in learning a variety of PDE's, e.g., \cite{patel2018MOR_Operator,patel2021MOR_Operator2,li2021FNO,Lu2021deeponet,rahman2023uno,Tapas2023wavelet}.


Neural operators utilizing the Fourier transform, e.g., MOR-Physics \cite{patel2018MOR_Operator} and Fourier Neural Operator (FNO) \cite{li2021FNO} are particularly attractive due to the simplicity of parametrization and computational efficiency. However, the Fourier transform exhibits Gibbs phenomenon when attempting to model discontinuities and cannot handle non-periodic boundary conditions. Fourier Neural Operators (FNOs) \cite{li2021FNO} tries to remedy this by adding ``bias functions'', \(Wv(x)\) that operate along the Fourier convolution layers. However, \cite{lu2022FNOvsDeepONet} has shown that this does not fully alleviate the weakness that FNO has around non-periodic boundary conditions and discontinuities, e.g., shock waves. Another neural operator, NORM \cite{Chen2024riemannian}, generalizes the modal strategy to Riemannian manifolds, but requires precomputation of the eigenfunctions of the Laplace-Beltrami operator and lacks a fast transform. \citep{Li2023deform}, learns in addition to an FNO, a deformation from a nontrivial domain to a periodic domain but is ill-suited for higher dimensional domains \cite{Chen2024riemannian}.


Instead, we address model discontinuities and boundary conditions by proposing a mixture of experts model, where the weighting of experts is determined locally across space via a partition of unity (POU). This mechanism enables the model to choose different experts on each side of a discontinuity or non-periodic transition, e.g. zero velocity at the walls for channel flow.  We apply a POU-Net \cite{lee2021POU_net, shazeer2017MoE_Sparse}  with a set of MOR-Physics Operators \cite{patel2018MOR_Operator} to better learn boundary conditions, introducing POU-MOR-Physics. In addition to being well suited to nontrivial boundary conditions, our method interpretably divides the domains for the experts. To demonstrate the method, we simultaneously learn the boundary conditions and a LES model for channel flow using the Johns Hopkins Turbulence Database (JHTDB) \cite{li2008JHTDB,graham2016JHTDB_channel,perlman2007JHTDB}.


We also take measures to accommodate the limited high-fidelity JHTDB DNS dataset. We embed a forward-Euler PDE solver in our model for training and implement uncertainty quantification (UQ) to provide a range of predictions outside the training data. The integration of the PDE solver with the learned correction serves a purpose similar to data augmentation: enabling higher-quality predictions despite the limited data. The UQ is implemented with Mean-Field Variational Inference (MFVI) \cite{blei2017variational}, which enables (1) the identification of the support of the dataset at prediction time (and hence detection of out-of-distribution (OOD) queries), and (2) the modeling of the notoriously high aleatoric uncertainty that are inherent to turbulence.

We leverage our neural operator to provide the missing physics for the subgrid scale dynamics in turbulent channel flow, i.e., LES closure modeling. To the best of our knowledge, the current state of the art for LES modeling of wall bounded turbulence with operator learning was presented by \citep{wang2024prediction}. The authors used an UNet enhanced FNO to learn a model for flows up to Re=590. In comparison, with the novel methods presented in this paper, our LES model incorporates UQ, \textit{a priori} known physics, models flows up to Re=1000, and is more interpretable due to Mixture-of-Expert style partitioning.

\subsection{Contributions}

In this work we,

\begin{itemize}
    \item Describe a connection between volume penalized numerical methods and POU-Nets
    \item Leverage the connection to develop a novel operator learning strategy capable of learning identifying multi-physics systems with non-trivial boundary conditions.
    \item Demonstrate the method in simple 2D operator learning problems.
    \item Advance the State-of-the-Art relative to \cite{wang2024prediction}, by accurately modeling Re=1000 3d wall bounded turbulence via neural operators and quantifying uncertainties.
\end{itemize}

\section{Methods}\label{methods}

Given pairs of functions as data, 
\begin{equation*}
    D = \left\{(u_i,v_i) | i\in {1,\hdots,n}, u_i \in U, v_i \in V\right\},
\end{equation*} 
where $U$ and $V$ are two Banach spaces, we seek an operator, $\mathcal{P}: U \rightarrow V$. The spaces $U$ and $V$ are sets of functions defined on $u_i: X \rightarrow \mathbb{R}^m$ and $v_i: Y\rightarrow \mathbb{R}^p$, where $X$ and $Y$ are compact subsets of $\mathbb{R}^{d_1}$ and $\mathbb{R}^{d_2}$, respectively.  Given our target application, we will consider $X=Y$ and therefore $d_1=d_2=d$. Operator learning introduces a parametrization for the unknown operator, $\mathcal{P}:U\times \Phi \rightarrow V$, where $\phi \in \Phi$ are parameters. As we demonstrate with learning an LES closure model in Section \ref{sec:les}, models can also incorporate \textit{a priori} known physics.

Given the parametrization, we can learn a deterministic model by solving the minimization problem,
\begin{equation} \label{eq:opt1}
    \phi = \underset{\hat\phi}{\mathrm{argmin}}\ L(D, \mathcal{P} (\cdot,\hat \phi)),
\end{equation}
where $L$ is a loss function, e.g., least squares. 
Alternatively, we can learn a probabilistic model by solving the optimization problem,
\begin{equation} \label{eq:opt2}
    Q = \underset{\hat Q}{\mathrm{argmin}}\ L'(D, \mathcal{P},\hat{Q}),
\end{equation}
where $Q$ is a variational distribution over $\Phi$ and $L'$ is a loss function, e.g., the negative of evidence lower bound (ELBO). 
The recovered operator must respect boundary conditions, which for neural operators leveraging the Fourier transform is a nontrivial task \cite{Li2023deform}.


One approach, volume penalization \cite{brown2014characteristic,kadoch2012volume}, is an embedded boundary method that integrates a variety of PDEs with complex boundary conditions. Volume penalization partitions a simple domain in two, where one of the subdomains is $X$, and applies forcing in each subdomain such that the solution to the new PDE, when restricted to $X$, approximates the solution to the original PDE. More generally, different kinds of physics may operate in different regions of space, and must be modeled as a collection of PDEs in disjoint domains, e.g., fluid-structure interaction \cite{engels2015numerical}. For these multi-physics problems, volume penalization partitions the extended domain accordingly with the appropriate forcing. In this work, we focus on Fourier pseudo-spectral methods which, while simple, efficient, and accurate,  only solve PDEs on periodic domains with Cartesian meshes in their base form. Applied to Fourier pseudo-spectral methods \cite{kolomenskiy2009fourier}, volume penalization retains the simplicity of the original discretization while expanding its reach to nontrivial problems. 

Our approach identifies physical systems with complex boundary conditions or multiphysics systems by learning the requisite partitions and forcing in a volume penalized Fourier pseudo-spectral scheme. We approach this task with the machine learning technique, mixture of experts, where experts learn their operations in subdomains of the problem space partitioned by learnable gating functions. 

Taking inspiration from volume penalization and mixture of experts, we parameterize $\mathcal{P}$ using a mixture of neural operators where each neural operator is a parameterized Fourier pseudo-spectral operator. This composite neural operator, which we refer to as POU-MOR-Physics, however, relies on the Fourier transform, so is designed for smooth functions with periodic domains. Ideally, a smooth extension for $u_i$ to this periodic domain is constructed to avoid Gibbs phenomena originating from the interface between $X$ and the periodic domain. We first discuss a strategy for constructing a smooth extension of $u_i$ in the next section that is compatible with the neural operator in Section \ref{sec:model}.



\subsection{Feature engineering -- Smooth extension of functions to periodic domains}\label{sec:extension}

Our neural operator, $\mathcal{P}$, discussed in Section~\ref{sec:model}, is parameterized via the Fourier transform and therefore is only well-defined for periodic functions. However, $U$ and $V$ are functions on the torus, $\mathbb{T}^d$, so we embed $X$ in $\mathbb{T}^d$, and must supply extensions for our functions in the new domain.  Our complete prediction mechanism, including the neural operator, restriction, and extension, takes the form,
\begin{equation}
    \mathcal{P}_e: u \overset{\mathcal{E}}{\mapsto} u_e \overset{{\mathcal{P}}}{\mapsto} v_e \overset{\mathcal{R}}{\mapsto} v.
\end{equation}
where $\mathcal{E}$ is an extension and $\mathcal{R}$ is a restriction; we have suppressed dependence on the parameters, $\phi$,

Since we only compute losses in $X$, we do not choose an extension for the output functions and allow our neural operators to produce actions (i.e. $v \in V$) that behave arbitrarily in the complement, $\mathbb{T}^d \setminus X$. Our procedure provides a smooth extension of input functions to the torus. Input functions with constant traces on the boundary are extended simply by setting the function value in $\mathbb{T}^d \setminus X$ to the constant value. For other functions, a similar approach would lead to discontinuities, which induce Gibbs phenomena due to the Fourier transforms in our parametrization.
In these cases, we solve a constrained optimization problem that minimizes the $H^1$ semi-norm of the extended function such that it matches the original function with domain, $X$.
\begin{equation}
\begin{matrix}
    \min_{u_e} & \int \nabla u_e \cdot \nabla u_e  dx \\
    \mathrm{s.t.}  & \mathcal{R} u_e = u \hfill
\end{matrix}
\end{equation}
The Lagrangian \cite{bertsekas2014constrained} for this optimization problem is,
\begin{equation}
    \mathcal{L} = \int \nabla u_e \cdot \nabla u_e  dx + (\lambda, \mathcal{R} u_e - u)_{L_2(\mathcal{U})}
\end{equation}
where the Lagrange multiplier, $\lambda \in {U}$. Using the Plancherel theorem, we rewrite the Lagrangian as,
\begin{equation}
    \mathcal{L} = \sum_\kappa (\kappa \mathcal{F} u_e) \cdot (\kappa \mathcal{F} u_e)  + (\lambda, \mathcal{R} u_e - u)_{L_2(\mathcal{U})}
\end{equation}
Using the unitary Fourier transform, $\mathcal{F}^{-1} = \mathcal{F}^*$, the first order conditions give a linear system,
\begin{equation}\label{eq:smooth_ext}
\begin{aligned}
\nabla_{[u_e,\lambda]^T} \mathcal{L} =0 \Rightarrow
    &\begin{bmatrix}
        \mathcal{F}^{-1} {\kappa \cdot \kappa} \mathcal{F} & \mathcal{R}^T \\
        \mathcal{R} & 0
    \end{bmatrix}
        \begin{bmatrix}
        u_e \\
        \lambda
    \end{bmatrix}
    =
        \begin{bmatrix}
        0 \\
        u
    \end{bmatrix}\\
    &=A q = b
\end{aligned}
\end{equation}
where $\mathcal{R}^T$ is given by $(\mathcal{R}u_e,u)_X=(u_e,\mathcal{R}^Tu)_{\mathbb{T}^d}$. We efficiently solve the normal equations for this system using matrix free operations in a conjugate gradient (CG) solver. 

\begin{figure}
    \centering
    \includegraphics[width=2in]{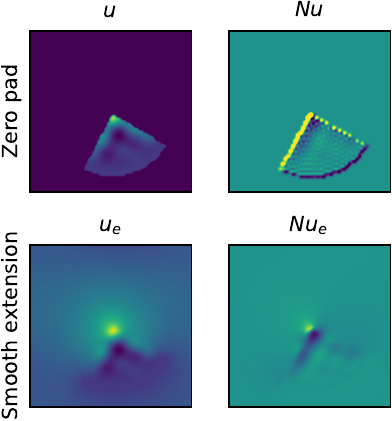}
    \caption{The smooth extension prevents Gibbs oscillation for Burgers action on a function originally defined on a quarter disk. \textit{(Top right)} Simple extension of input function to torus via zero padding. \textit{(Bottom right)} Smooth extension of input function to torus. \textit{(Top left)} Burgers action on simple extension. \textit{(Bottom left)} Burgers action on smooth extension.}
    \label{fig:smooth_extention}
\end{figure}

To illustrate this method, Figure \ref{fig:smooth_extention} compares the smooth extension of an input function with an extension that sets the function values to zero outside of the domain and compares the action of the Burgers operator, $v = u\cdot \nabla u$, on the two extensions. The discontinuity in the simple extension leads to an oscillatory action while the action from the smooth extension is less oscillatory. This test was performed using input function data generated as per Section \ref{sec:wedge}.

For the remainder, we will drop the subscript, $e$, and identify $u_e$ with $u$, $v_e$ with $v$, and $\mathcal{P}_e$ with $\mathcal{P}$.

\subsection{Model Design -- Partition of Unity network with MOR-Physics neural operators}\label{sec:model}

We now construct a neural operator, POU-MOR-Physics, for functions with domain, $\mathbb{T}^d$. 
Our parameterization is a spatially conditioned POU-network composed of a gating network and neural operator experts:
\begin{equation}\label{eq:pou_model}
    (\mathcal{P} (u;\phi)) (x) = \sum_{i=1}^I G_i(x;\phi_{G_i}) (\mathcal{N}_i(u;\phi_{\mathcal{N}_i}))  (x),
\end{equation}
where $G$ forms a partition of unity, i.e. $\forall x \in \mathbb{T}^d$,
\begin{equation}
    \sum_i^I G_i (x;\phi_{G_i}) = 1 \quad \text{and} \quad G_i(x;\phi_{G_i}) \ge 0,
\end{equation}
and where $\phi$ are the combined set of parameters in the full model and $\phi_*$ are parameters for the various subcomponents of the model. Where the context is clear, we will suppress in our notation the dependence on the parameters. We describe each of the components of \(\mathcal{P}\) -- the neural operators $\mathcal{N}_i$, $i=1,\dots,I$,  the gating network, $G$, and the time-dependent autoregressive strategy -- in turn, which then situates the model to be used for mean-field variational inference (MFVI).

\subsubsection{MOR-Physics Operator}

The neural operators in Equation \eqref{eq:pou_model}, $\mathcal{N}_i$, are modified versions of the MOR-Physics Operator presented in \cite{patel2018MOR_Operator,patel2021MOR_Operator2}. For convenience, we drop the subscript $i$, as we define each $\mathcal{N}_i$ operator similarly. We will exclusively parameterize operators with the approach detailed below. However, if we have \textit{a priori} knowledge about a system, we can use a predefined operator in addition to the learned ones. In our exemplars, we include the zero expert in our ensemble of experts, i.e., an operator that evaluates to zero for any input function.

 We compose $L$ MOR-Physics operators, $\mathcal{N}^{(l)}$, $l = 1:L$, thereby introducing latent functions, $v^{(l)}:\mathbb{T}^{d} \rightarrow \mathbb{R}^{m^{(l)}}$, after the action of each operator, 
\begin{equation}
    v^{(l+1)} = \mathcal{N}^{(l+1)}(v^{(l)}) =\mathcal{F}^{-1}(g^{(l+1)}\cdot \mathcal{F}(h^{(l+1)} \circ v^{(l)})),
\label{eq:MOR_layer}
\end{equation}
where $\mathcal{F}$ is the Fourier transform, $h^{(l)}$ is a learned local activation, implemented with a neural network, and $g^{(l)}$ is a weighting function in Fourier frequency space.
\begin{figure*}
    \centering
    \includegraphics[width=1.0\linewidth]{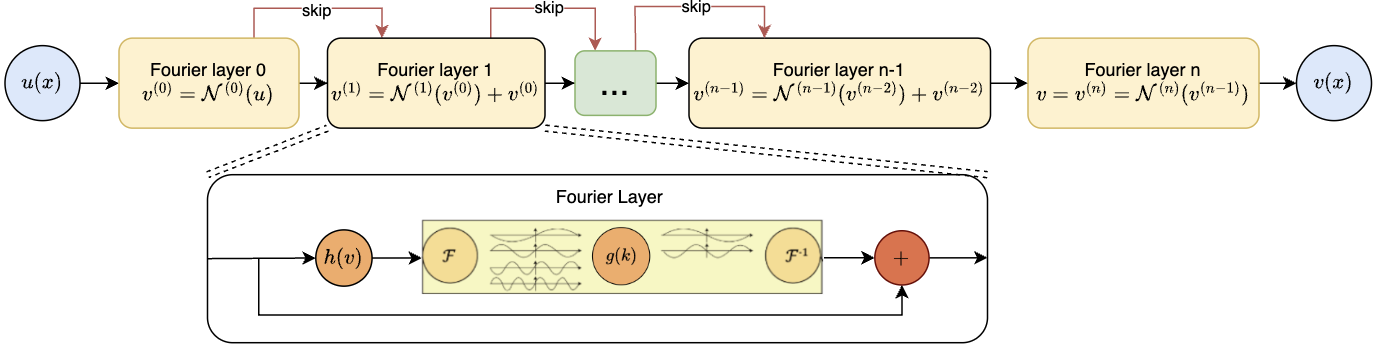}
    \caption{MOR Operator diagram, showing architecture of each expert operator $\mathcal{N}_i$. Black arrows denote function composition.}
    \label{fig:MOR_Operator}
\end{figure*}
The multiplication operation, $\cdot$, in Equation \eqref{eq:MOR_layer} is both the Hadamard product and a matrix multiplication, i.e., at every wavenumber, $k$, we have a matrix-vector product between \(g^{(l+1)}(k) \in \mathbb{C}^{m^{(l+1)} \times  m^{(l)}}\) and \(\mathcal{F}(h^{(l+1)} \circ v^{(l)})(k) \in \mathbb{C}^{m^{(l)}}\). This operation is comparable to linear operations across channels in a convolutional neural network (CNN).

To build these operators, we discretize the domains on Cartesian meshes and replace the abstract linear operations with the appropriate matrix operations. Depending on the problem, we implement \(g^{(l)}(k)\) either as a neural network or a parameterized tensor. In either case, we truncate the higher modes of either the \(g^{(l)}(k)\) or \(\mathcal{F}(v^{(l)})(k)\) tensors (which ever has more in a given dimension) so that their shapes are made compatible. This effectively results in a low pass filter; more details can be seen in \cite{patel2018MOR_Operator}.

All hidden layers are appended with a skip connection. Ultimately, each $\mathcal{N}_i$ is then built via composition of several layers $\mathcal{N}^{(l)}$, so that
$$\mathcal{N}_i(u) := (\mathcal{N}^{(L)} \circ \mathcal{N}^{(L-1)} \circ \dots \circ \mathcal{N}^{(0)})(u).$$
and $v^{(L)} = v$ provides is the action of the composite MOR-Physics operator.
This architecture is sketched in Figure \ref{fig:MOR_Operator}, mapping from an input $u$ to a target $v$ through the intermediate layers $v^{(l)}$.

\subsubsection{Gating Network}\label{sec:gates}

\begin{figure}[htbp!]
    \centering
    \includegraphics[width=1.0\linewidth]{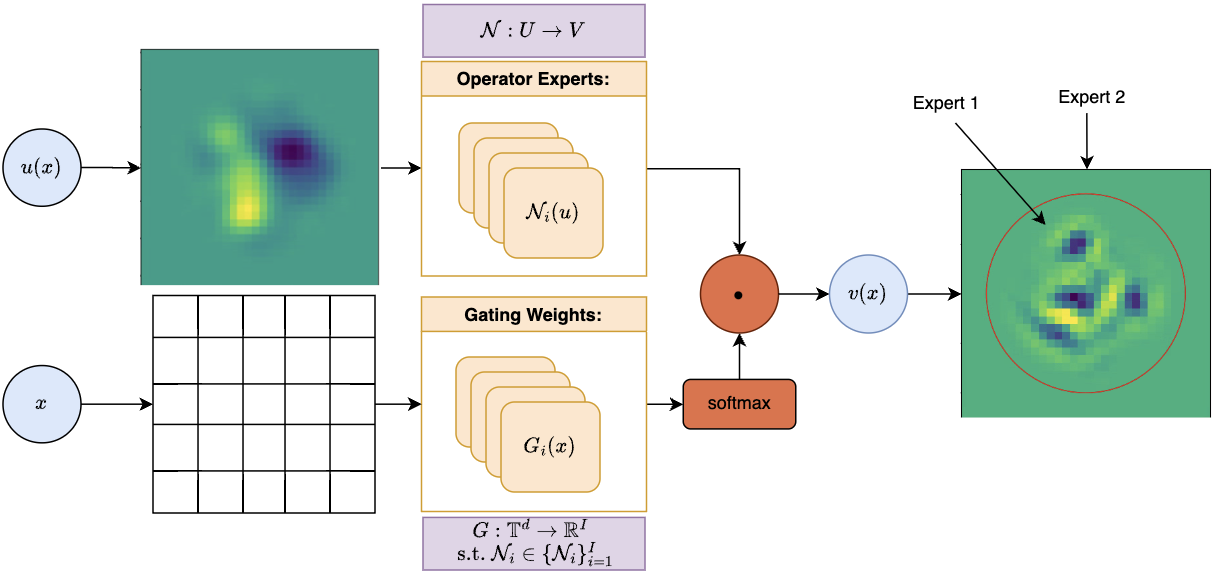}
    \caption{Mixture of Experts model, where a weighted sum of gating weights \(G_i(x)\) is applied to expert outputs \(\mathcal{N}_i(u)\). The gating weights are spatially localized, depending only on $x$ and not $u$. }
    \label{fig:Mixture_of_Experts}
\end{figure}

The gating network $G: \mathbb{T}^d \rightarrow \mathbb{R}^I$ is built from a neural network, taking the coordinates as inputs.
Its output is transformed via softmax into coefficients to compute a convex combination of neural experts, \(\mathcal{N}_i\), at location \(x \in \mathbb{T}^d\). The Gating network does \textit{not} take \(u(x)\) as input, it only uses the location \(x\) which is sufficient to partition the space for different experts. See Figure \ref{fig:Mixture_of_Experts} for a schematic of the gating function.



Since we rely on the Fourier transform for our operator parameterization, we use a smooth mapping, $\mathbb{T}^d \rightarrow \mathbb{R}^{2d}$, to provide the input to the gating network. The domain, $\mathbb{T}^d$, is parameterized by $d$ angles, $\theta_i$ and mapped to a vector in $\mathbb{R}^{2d}$ as, 
\begin{equation}
    [\sin(\theta_1), \hdots, \sin(\theta_d),\cos(\theta_1), \hdots, \cos(\theta_d)]^T.
\end{equation}
This yields more consistently interpretable expert partitions; In our 2D exemplar discussed in Section \ref{sec:2d_data} we  found this approach to produce symmetric partitions that conform to the problem specification, while the simpler approach using the angles as coordinates led to asymmetric partitions.

Although we describe a neural network gating function in this section, $\textit{a priori}$ known gates can replace the network, e.g., when a PDE domain is already well characterized.


\subsubsection{Autoregressive POU-MOR-Physics model}\label{sec:autoregressive}

Autoregressive models are well-suited for learning spatiotemporal dynamics. We can specialize the model in \eqref{eq:pou_model} to learn an update operator representing the evolution of the system over a small period of time, $\Delta t$, by letting $\mathcal{P}:\mathcal{U}\rightarrow \mathcal{U}$ and,
\begin{equation}
\begin{split}
    u(\cdot,(n+1)\Delta t) & = \mathcal{P} u(\cdot,t)\\
    = u_{n+1} &=  \mathcal{P} u_n.
\end{split}
\end{equation}
In this context, the operator $\mathcal{P}$ can be composed with itself to predict the system at discrete times, $u_{n+p} =  \mathcal{P}^p u_n$. In many cases, parts of a model are \textit{a priori} known and we have a PDE with an unknown operator,
\begin{equation}
    \begin{aligned}\label{eq:autoreg1}
        \partial_t u = \mathcal{M} u + \tilde{\mathcal{P}}u
    \end{aligned}
\end{equation}
where $\mathcal{M}$ is known and $\tilde{\mathcal{P}}$ is unknown. A first order in time operator splitting allows for the update,
\begin{equation}\label{eq:autoreg2}
        u_{n+p} = \left[ \mathcal{P}(I+\Delta t \mathcal{M})\right]^pu_n = U^pu_n\\
\end{equation}
where $\mathcal{P}$ is a new unknown operator that provides the same effect as $\tilde{\mathcal{P}}$. Since  the Euler update operator, $(I+\Delta t)\mathcal{M}$, is a common time integrator for PDEs, we will refer to it as the PDE solver for the remainder of this work.
We will demonstrate an autoregressive model in Section \ref{sec:les} by learning an LES closure model for wall bounded turbulent flow.

 We introduce the operator, $\overline{\mathcal{U}}^p$, to give a time series prediction of $p$ time steps from an initial field, and denote its action on an initial condition, $\overline{\mathcal{U}}^p u_n = u_{[n,n+p]}$, where the subscript indicates a time-series beginning from timestep $n$ and ending with timestep $n+p$.

 Given a time-series of functional data, $D = \{\tilde{u}_n\}^{n=0,\hdots,N}$, an autoregressive model can be found by solving an optimization problem, i.e., \eqref{eq:opt1} or \eqref{eq:opt2}.



\subsection{Mean-Field Variational Inference (MFVI)}\label{VI}

\begin{figure}[htpb!]
    \centering
    \includegraphics[width=1.0\linewidth]{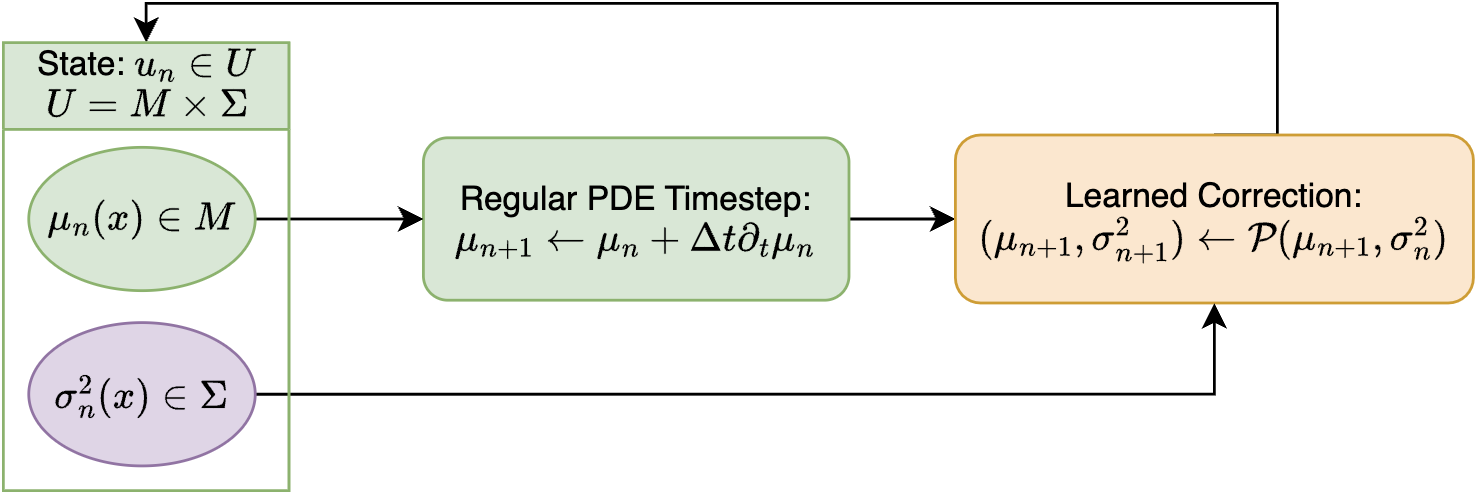}
    \caption{Flow of Uncertainty through Learned PDE solver model. The PDE solver operates on the mean field, $\mu_n \in M$, while the uncertainty field, $\sigma_n^2 \in \Sigma$ is entirely modeled and updated by the Neural Operator.}
    \label{fig:Learned_Correction_UQ}
\end{figure}

The auto-regressive nature of our Bayesian MFVI model (in Section~\ref{sec:les}) necessarily changes our otherwise standard implementation of MFVI. The details of the major difference are highlighted in Figure \ref{fig:Learned_Correction_UQ}.

To build MFVI into our neural operator, we assume independent Gaussian variational posteriors for each model parameter, exploiting the conventional reparameterization trick \cite{kingma2013VAEs} for normally distributed weights; see Appendix \ref{Complex_Reparameterization} for how it applies to complex parameters. Further details on our VI method can be found in \cite{blundell2015weight}.

Following \cite{blundell2015weight}'s convention we treat the VI model as a probabilistic model, in the sense that it directly predicts the mean and variance of Gaussian likelihoods, that is \(\mathcal{P}: U \rightarrow U,\) where $U=M\times \Sigma$, such that the mean and variance  are given by $u=(\mu,\sigma^2)\in U$. Only the $\mu$ fields are integrated by the PDE solver as described in Figure \ref{fig:Learned_Correction_UQ}. Given a time-series of data, $\tilde{v}$, we learn a Bayesian model using variational inference by optimizing the ELBO,
\begin{equation}\label{eq:opt_vi}
\begin{aligned}
    \max_{q}\ (\log p\left(\tilde{v}_{[0,N]}|\overline{\mathcal{U}}^P\tilde{u}_0 \right) - D_{KL}(q || p_0))
\end{aligned}
\end{equation}
where $q$ is a variational distribution over $\Theta$, $p_0$ is a prior distribution over $\Theta$,  $\log p$ is a Gaussian log likelihood with mean and variance provided by $u$. 
Equation \eqref{eq:opt_vi} is not computationally tractable, so we use a strided sliding window strategy and obtain,
\begin{equation}\label{eq:opt_vi2}
\begin{aligned}
    \max_{q} \sum_{m=0:N:P}(\log p\left(\tilde{v}_{[m,m+P]}|\overline{\mathcal{U}}^P\tilde{u}_m \right) - D_{KL}(q || p_0))
\end{aligned}
\end{equation}

where $\tilde{u}_{[n,n+P]}$ is the subset of data from timestep $n$ to $n+P$ and  $P$ is a hyperparameter for the number of autoregressive steps used during training. The notation, $0:N:P$, borrows from Python's array slicing syntax. To initialize the means and variances, we set, \(\forall m, \tilde u_m=(\tilde v_m, \tilde \sigma_m^2=0)\). We approximate the sum via stochastic mini-batch optimization. Overlapping sliding windows would constitute data-augmentation, which has been shown to cause the cold posterior effect (CPE) \cite{izmailov2021bayesian}, therefore we avoid using them.

With MFVI, the resulting model outputs Gaussian distribution predictions by applying a weighted sum of the experts' output tensors, c.f. Equation \eqref{eq:pou_model}, where now these output tensors contain both \(\mu\) and \(\sigma^2\) channels for each output feature.

The variance $\sigma^2$ is parameterized by learning a parameter $\rho$ and then computing
\begin{equation}
    \sigma^2(\rho) = (\log(1 + e^{\rho}))^2
    \label{eq:sigma_squared_parameterization}
\end{equation}
The squared softplus positivity constraint in Equation \eqref{eq:sigma_squared_parameterization} provides both the model predictions of $\sigma^2$ and the parameterization of the variances in the Gaussian variational distributions needed for VI. While enforcing positivity via squaring \(\sigma^2=\rho^2\) would be simpler, we believe our parametrization promotes easier learning.

\section{Numerical demonstrations}

We demonstrate our method with three numerical examples. This section describes the problem formulations and our results. In the first example, we learn a nonlinear operator for functions on a disk. In second, we learn the solution operator to a nonlinear poisson equation with mixed boundary condition. In the third, we learn an LES model for turbulent channel flow.




\subsection{Recovery of a nonlinear operator on the unit disk}\label{sec:2d_data}

Our first example demonstrates POU-MOR-Physics is capable of learning local operators in nontrivial domains. We consider synthetically generated pairs of functions, ($u_i,v_i$), on the unit disk that vanish on the boundary, $\Omega = \left\{(x,y):x^2+y^2\leq 1\right\}$. We first generate a function $\hat{u}_i$ by sampling a Gaussian process. We construct a regular grid of points over a box in which $\mathcal{X}$ is embedded, and compute the point evaluations of the function $\hat{u}$ on the grid by sampling from a spatial Gaussian process with a fixed kernel $K$; the function $\hat{v}_i$ is then manufactured by a second-order finite difference approximation of the Laplacian of $\hat{u}_i^2$. We retain the function evaluations in the disk as our data, ($u_i,v_i$).
Our loss function is then the squared error, 
\begin{equation}
    L = \sum_i || \mathcal{R}{\mathcal{P}}(\mathcal{E}{{u}}_i; \theta) - {v}_i||_{\ell_2}^2
\end{equation}

\begin{figure}[htpb!]
    \centering
        \includegraphics[width=0.3\linewidth]{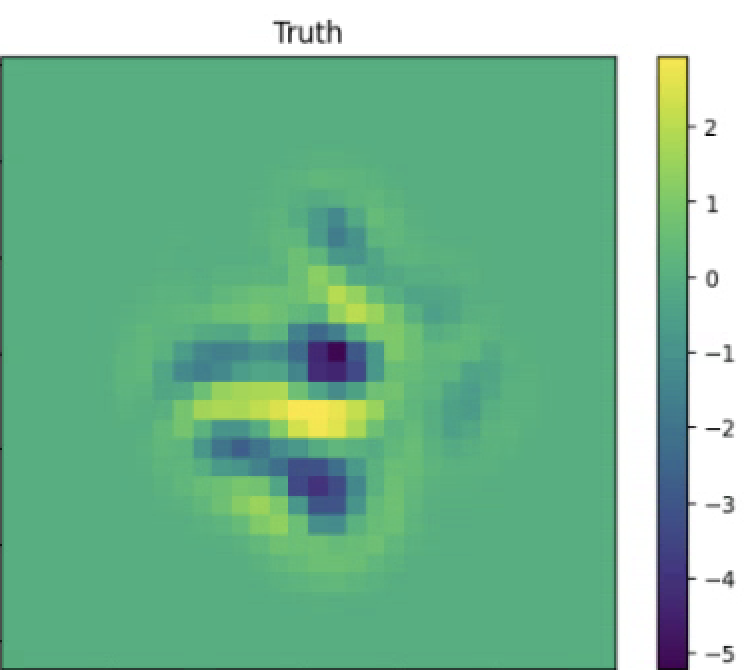}
        \hspace{.1in}  
        \includegraphics[width=0.3\linewidth]{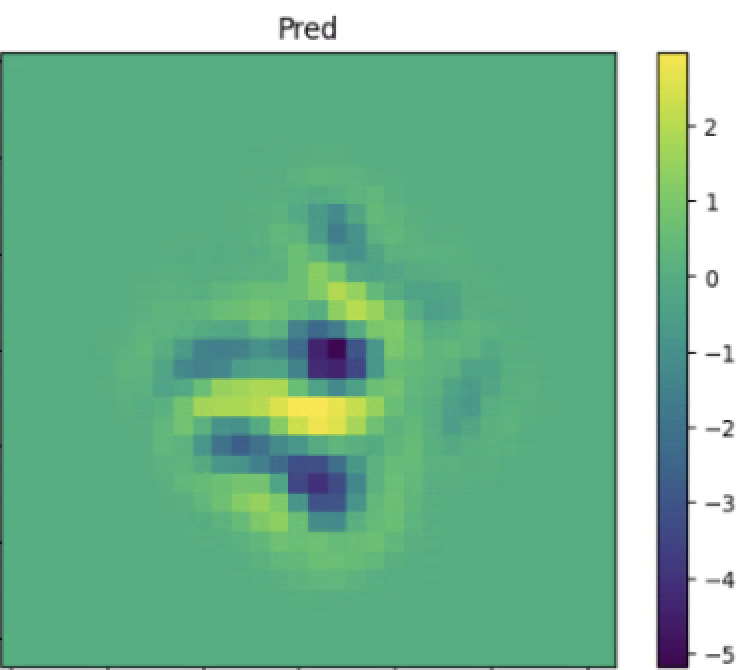}
        \hfill \\
        \includegraphics[width=0.3\linewidth]{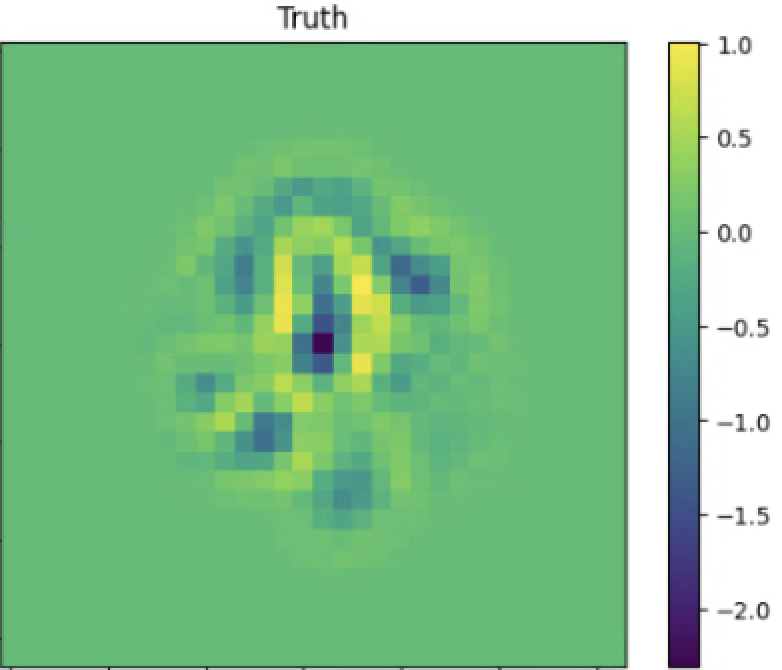}
        \hspace{.1in}  
        \includegraphics[width=0.3\linewidth]{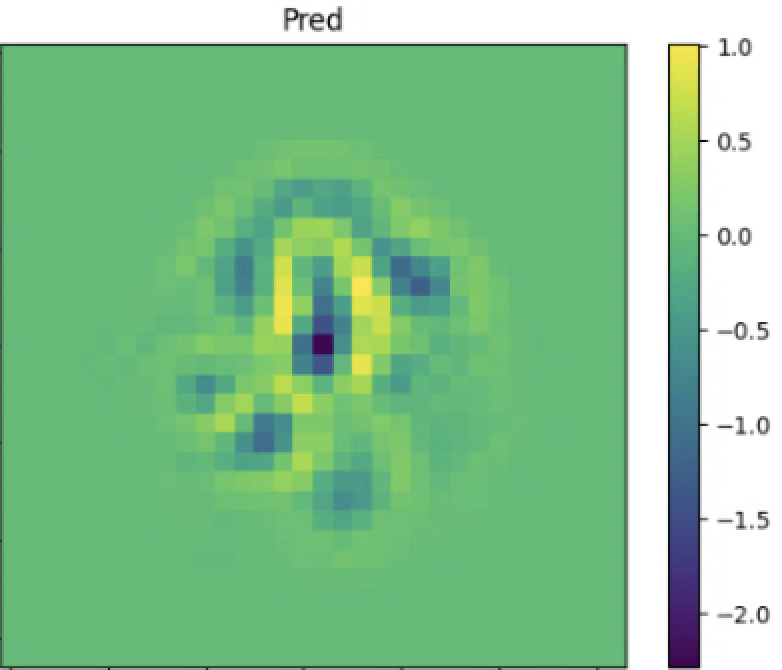}
    \caption{(Left) Test data and (Right) prediction 2D synthetic exemplar. Top and Bottom are results from different test data samples.}
    \label{fig:Truth_Pred2d}
\end{figure}

\begin{figure}[H]
    \centering
    \includegraphics[width=0.7\linewidth]{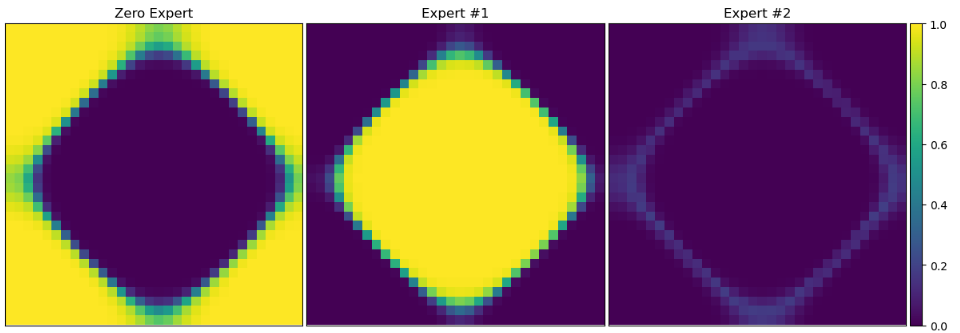}
    \caption{Learned 2D expert partitions for 2D synthetic data. We clearly see a ``zero expert'' learned to capture boundary behavior.}
    \label{fig:2d_expert_partitions}
\end{figure}

On this synthetic 2D data problem, the model successfully recovers the true solution; the model's outputs are visualized in Figure~\ref{fig:Truth_Pred2d}. We achieve a validation \(R^2>99.999\%\); as seen in Figure~\ref{fig:2d_expert_partitions}, we successfully assign a  ``Zero Expert'' \(Z(a)=0, Z: A \rightarrow U\) to the boundaries as is intended, allowing the recovery of the true boundary conditions for this problem.

\subsection{Solution operator for nonlinear Poisson problem on quarter disk}\label{sec:wedge}

\begin{figure}
    \centering
    \includegraphics[width=2in]{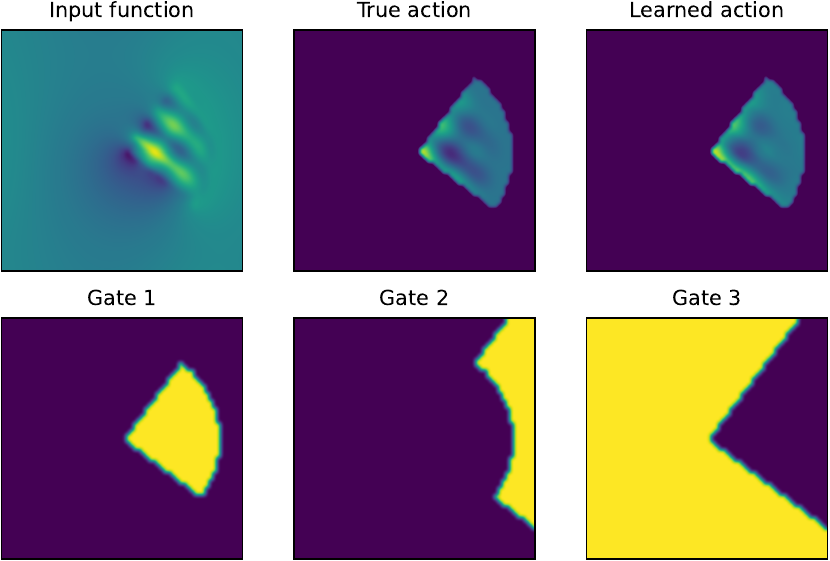}
    \caption{POU-MOR operator learns the solution operator for a nonlinear Poisson equation with nontrivial and mixed boundary conditions.}
    \label{fig:poisson}
\end{figure}

In this section, we demonstrate POU-MOR-Physics is capable of learning solution operators for nonlinear PDEs with complex domains and mixed boundary conditions. We begin by generating pairs of data, $\left(u_i,v_i\right)$, that solve a nonlinear Poisson equation and learn the solution operator. Dropping the subscripts, we first  sample the functions, $\left(\hat{u},\hat{v}\right)$,
\begin{equation}
\begin{aligned}
    &\psi(x) = \left\{ \begin{matrix}1-||x||_2^2 & if\ ||x||_2^2<1\\
    0 & else\end{matrix} \right.\\
    &\hat{v}(x) = \psi(x)\sum_{m=1}^{10} \cos (f_{m,1} x_1) \cos (f_{m,2} x_2) \\
    &\hat{u} = \nabla \cdot \tanh(\nabla u)
\end{aligned}
\end{equation}
where $f$ are random frequencies, $f \sim U[0,10]$. Restricting the domains of $\hat{u}$ and $\hat{v}$, to the quarter disk, $\{x: ||x||_2^2\leq 1, x\geq 0\}$, we obtain the data, $\left\{\overline{u},\overline{v}\right\}$, which solves the nonlinear Poisson equation,
\begin{equation}
    \begin{matrix}
        \nabla \cdot \tanh(\nabla \overline{v}) = u & x\in \Omega\\
        \nabla \overline{v} \cdot n =0 & \begin{aligned}
            \{(x_1,0): x_1 \in [0,1]\} \cup \\\{(0,x_2): x_2 \in [0,1]\}
        \end{aligned} \\
        \overline{v} = 1 & \{ x: 0\leq\arctan \frac{x_2}{x_1}\leq\frac{\pi}{2}\}\
    \end{matrix}
\end{equation}
we next rotate the domains of $\overline{u}$ and $\overline{v}$ by a random rotation matrix and obtain our data, ${u}$ and ${v}$. We obtain 10000 samples of these input/output function pairs and train a POU-MOR-Physics operator using the least squares loss. For this example, we use predefined gates that conform to the problem specification instead of a gating network (see Section \ref{sec:gates}). Figure~\ref{fig:poisson} compares the learned action on test data to the true action. We obtain approximately 1\% relative RMSE for this action compared to the true action.

\subsection{Large Eddy Simulation Modeling}\label{sec:les}

Our target application is extracting an LES model from DNS provided by the JHTDB dataset \cite{graham2016JHTDB_channel}. The simulation data is obtained from the  Re=1000 channel flow problem with no-slip boundary conditions (BCs) on the top and bottom of the flow, and periodic BCs on the left, right, front and back.

The JHTDB channel flow problem \cite{graham2016JHTDB_channel}, characterized by three spatial dimensions plus a recursive time dimension, presents significant computational challenges that demand careful resource management. To mitigate the substantial memory requirements, we utilized the real Fast Fourier Transform (rFFT) within the MOR-Physics Operator \cite{patel2021MOR_Operator2}, effectively conserving memory without compromising performance.



We subsample the DNS data spatially (see Table~\ref{tab:post_processing_parameters}) and keep the full resolution in the time dimension, motivated by the need to have a larger training dataset. The spatial sub-sampling is performed after applying a box filter to the DNS data, ensuring the sub-sampled grid is representative of the whole DNS field. 

From this data, we seek to obtain an autoregressive (Section \ref{sec:autoregressive}), MFVI (Section \ref{VI}) model that incorporates physics by leveraging the standard LES formulation (Appendix \ref{sec:les_apriori}). While the JHTDB dataset only simulates the flow for one channel flow through time, $t=T$, we use our model to extrapolate the flow to long times, $t=10T$, and evaluate the predictions and uncertainties from our model.



\begin{table}[H]
    \centering
    \begin{tabular}{|l|c|} \hline 
        
        \textbf{Parameter} & \textbf{Value} \\ \hline 
        
        Spatial Stride & \( s_{x} = s_{y} = s_{z} = 20 \) \\ \hline 
        
        Sub-Sampled Dimensions & \( 103 \times 26 \times 77 \) \\ \hline 
        
        Time Dimension & \( t = 4000 \) \\ \hline 
        
        Box Filter Dimension & \( b = 20 \) \\ \hline
        
    \end{tabular}
    \caption{Post Processing Parameters for JHTDB problem.}
    \label{tab:post_processing_parameters}
\end{table}

\subsubsection{Training details}

Due to the memory demands of this learning task, we required parallelization to train our LES model.
To handle the computational load efficiently, we employ 20 A100 GPUs alongside the Fully-Sharded Data Parallel (FSDP) strategy \cite{zhao2023FSDP}, thereby exploiting data-parallel training and necessitating a scaled learning rate that follows the batch size \cite{imagenet1hour2017}.


We used the linear scaling rule from \cite{imagenet1hour2017} and the "One Cycle" warm-up schedule from \cite{smith2019super}; without warm-up, the model parameters may change too rapidly at the outset \cite{imagenet1hour2017} for effective training.

Adhering to the linear scaling rule from \cite{imagenet1hour2017}, we set the batch size to \(\text{batch\_size} = n\) and the learning rate to \(\text{learning\_rate} = 1.25\times10^{-4}*n\), where $n$ represents the number of GPUs utilized. Although memory constraints necessitated a small per-GPU batch size of 1, we found that combining this approach with gradient clipping yielded effective training results.

To improve the stability of the learned subgrid dynamics operator, we employed several (i.e. 8) auto-regressive time steps during training, following methodologies similar to those proposed by Bengio et al. \cite{bengio2015scheduled} (but without the curriculum). Statistical auto-regressive time series models often suffer from exponentially growing errors because the model's flawed outputs feed back into its inputs, compounding errors over multiple time steps. By exposing the model to this process during training — taking several auto-regressive steps before each optimization step — the model learns to correct its own compounding errors, mitigating  problems at prediction time \cite{bengio2015scheduled}.


\subsubsection{LES model Results}

\begin{figure}[H]
    \centering
    \includegraphics[width=0.5\linewidth]{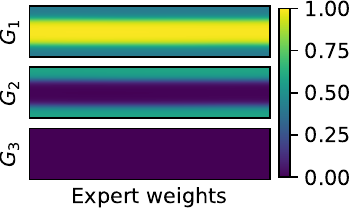}
    \caption{3D Expert Partitions for JHTDB dataset. $G_3$ is the zero expert, i.e., $\mathcal{N}_3 = 0$.}
    \label{fig:les_expert_partitions}
\end{figure}

In Figure \ref{fig:les_expert_partitions}, we find that the gating network is able to  partition the domain and find separate models for the bulk and boundary layers of the JHTDB channel data. We see spatial partitions that reflect the boundary layer, with different support for $\mathcal{N}_1$ vs. $\mathcal{N}_2$ Moreover, the transition from \(\mathcal{N}_1\) to \(\mathcal{N}_2\) is continuous roughly approximating the strength of the velocity at those points in the simulation. This demonstrates the model's ability to find multiple, spatially conditioned models.


\begin{figure}[H]
    \centering
    \includegraphics[width=\linewidth]{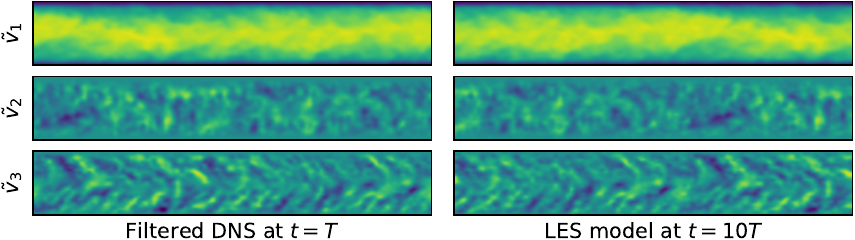}
    \caption{Fields from filtered \textit{(Left)} DNS at $t=T$ and learned  \textit{(Right)} Sample of LES model predictions after 10 channel flow-through times.}
    \label{fig:Qcrit}
\end{figure}

\begin{figure}[H]
    \centering
    \includegraphics[width=\linewidth]{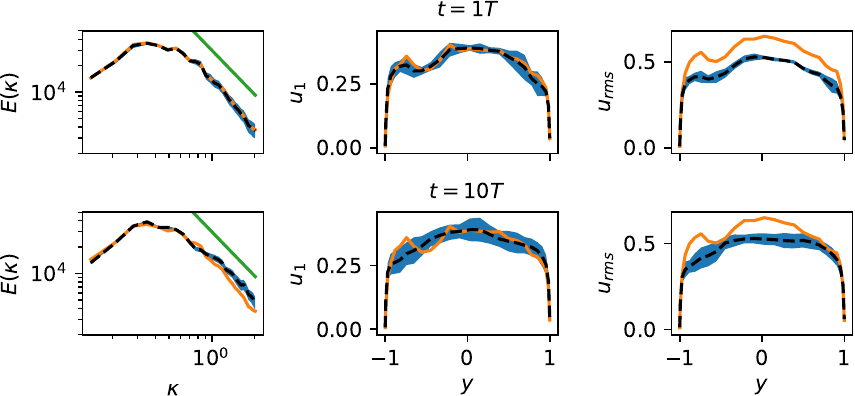}
    \caption{MFVI LES model captures \textit{(Left)} energy spectrum and \textit{(Center)} bulk velocity, and \textit{(Right)} RMS fluctuations. Posterior predictive samples are shown in black. The DNS results are shown in orange. Kolmogorov's 5/3 rule is shown in blue.}
    \label{fig:Energy_Spectrum}
\end{figure}

We use our LES model to evolve the system to 10 channel flow-through times. Since the JHTDB only includes one flow length's worth of data, we are predicting dynamics in an extrapolatory regime and use the VI model to provide predictions with error bars. Since the flow is at a statistical steady state, we should see the same mean statistics for our model as the DNS data. Figure \ref{fig:Energy_Spectrum} shows a statistical comparison between our model and the filtered DNS data. We see that the energy spectrum closely matches that of DNS and obey's Kologmorov's 3/5's rule (the error bars don't pass the threshold). In the middle figures we see close agreement also even at t=10T. However we also see the uncertainty is as expected increasing with time. Finally we see decent agreement with RMS velocity fluctuations in the right most figures, however there is still some room for improvement as matching RMS fluctuations \cite{Pope_2000} is more difficult.

We also were able to achieve $98.8\% R^2$ reproducing the simulation with a deterministic model; the very closely matching fields are shown in the Appendix \ref{sec:deteministic_les}; see, e.g. Figures \ref{fig:deterministic_x_velocity},\ref{fig:deterministic_y_velocity}, and \ref{fig:deterministic_z_velocity}. 


\subsection{Out-of-Distribution (OOD) Detection:}

\begin{figure}
    \centering
    \includegraphics[width=0.3\linewidth]{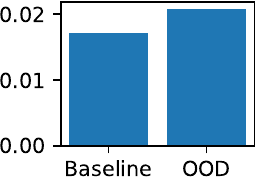}
    \hspace{.1in}
    \includegraphics[width=0.6\linewidth]{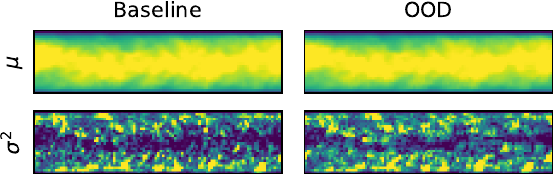}
    \caption{We test Bayesian VI model's OOD detection by comparing the uncertainty from a learned simulation with a in-distribution initial-condition (IC): JHTDB data with a box filter size of 20, to the uncertainty from an OOD IC: JHTDB data with no box filter. The model's uncertainty can clearly detect that the no-filter initial condition (IC) is OOD. Even though all it has to go on is only a single time-slice (the IC).}
    \label{fig:ood}
\end{figure}

In Figure \ref{fig:ood} we validate the Bayeisan model's UQ OOD detection ability. This makes us confident in using the UQ to validate the extrapolation predictions for which we do not have ground truth to compare.

\section{Conclusion}
In this work, we addressed the challenge of learning non-periodic boundary conditions and discontinuities in PDE simulations using a Mixture of Experts (MoE) approach, specifically applied to the MOR-Physics neural operator. The motivation for this work stems from the limitations of traditional Fourier-based PDE solvers, which struggle with non-periodic boundary conditions, such as those found in Navier-Stokes simulations. By integrating a forward Euler PDE solver and applying a correction operator at each timestep, we were able to enhance solver accuracy. Our methods were tested on both 2D synthetic data and LES for 3D channel flow, demonstrating promising results. The model was highly successful in the 2D case, where it flawlessly captured the zero boundary conditions. Furthermore, the Bayesian 3D channel flow problem yielded promising results with a close match between the energy spectra of the DNS and model predictions.




\section*{Acknowledgments}

Sandia National Laboratories is a multi-mission laboratory managed and operated by National Technology and Engineering Solutions of Sandia, LLC., a wholly owned subsidiary of Honeywell International, Inc., for the U.S. Department of Energy’s National Nuclear Security Administration under contract DE-NA0003525. This paper describes objective technical results and analysis. Any subjective views or opinions that might be expressed in the paper do not necessarily represent the views of the U.S. Department of Energy or the United States Government.

SAND Number: SAND2025-01380O

\bibliography{main.bib}

\begin{thebibliography}{30}
\providecommand{\natexlab}[1]{#1}
\providecommand{\url}[1]{\texttt{#1}}
\expandafter\ifx\csname urlstyle\endcsname\relax
  \providecommand{\doi}[1]{doi: #1}\else
  \providecommand{\doi}{doi: \begingroup \urlstyle{rm}\Url}\fi

\bibitem[Pope(2000)]{Pope_2000}
Stephen~B. Pope.
\newblock \emph{Turbulent Flows}.
\newblock Cambridge University Press, 2000.

\bibitem[Chen and Chen(1995)]{Chen1995}
Tianping Chen and Hong Chen.
\newblock Universal approximation to nonlinear operators by neural networks
  with arbitrary activation functions and its application to dynamical systems.
\newblock \emph{IEEE Transactions on Neural Networks}, 6\penalty0 (4):\penalty0
  911--917, 1995.
\newblock \doi{10.1109/72.392253}.

\bibitem[Patel and Desjardins(2018)]{patel2018MOR_Operator}
Ravi~G Patel and Olivier Desjardins.
\newblock Nonlinear integro-differential operator regression with neural
  networks.
\newblock \emph{arXiv preprint arXiv:1810.08552}, 2018.

\bibitem[Patel et~al.(2021)Patel, Trask, Wood, and Cyr]{patel2021MOR_Operator2}
Ravi~G. Patel, Nathaniel~A. Trask, Mitchell~A. Wood, and Eric~C. Cyr.
\newblock A physics-informed operator regression framework for extracting
  data-driven continuum models.
\newblock \emph{Computer Methods in Applied Mechanics and Engineering},
  373:\penalty0 113500, 2021.
\newblock ISSN 0045-7825.

\bibitem[Li et~al.(2021)Li, Kovachki, Azizzadenesheli, liu, Bhattacharya,
  Stuart, and Anandkumar]{li2021FNO}
Zongyi Li, Nikola~Borislavov Kovachki, Kamyar Azizzadenesheli, Burigede liu,
  Kaushik Bhattacharya, Andrew Stuart, and Anima Anandkumar.
\newblock Fourier neural operator for parametric partial differential
  equations.
\newblock In \emph{International Conference on Learning Representations}, 2021.

\bibitem[Lu et~al.(2021)Lu, Jin, Pang, Zhang, and Karniadakis]{Lu2021deeponet}
Lu~Lu, Pengzhan Jin, Guofei Pang, Zhongqiang Zhang, and George~Em Karniadakis.
\newblock Learning nonlinear operators via deeponet based on the universal
  approximation theorem of operators.
\newblock \emph{Nature Machine Intelligence}, 3\penalty0 (3):\penalty0
  218--229, 2021.
\newblock \doi{10.1038/s42256-021-00302-5}.

\bibitem[Rahman et~al.(2023)Rahman, Ross, and Azizzadenesheli]{rahman2023uno}
Md~Ashiqur Rahman, Zachary~E Ross, and Kamyar Azizzadenesheli.
\newblock U-{NO}: U-shaped neural operators.
\newblock \emph{Transactions on Machine Learning Research}, 2023.
\newblock ISSN 2835-8856.

\bibitem[Tripura and Chakraborty(2023)]{Tapas2023wavelet}
Tapas Tripura and Souvik Chakraborty.
\newblock Wavelet neural operator for solving parametric partial differential
  equations in computational mechanics problems.
\newblock \emph{Computer Methods in Applied Mechanics and Engineering},
  404:\penalty0 115783, 2023.
\newblock ISSN 0045-7825.
\newblock \doi{https://doi.org/10.1016/j.cma.2022.115783}.

\bibitem[Lu et~al.(2022)Lu, Meng, Cai, Mao, Goswami, Zhang, and
  Karniadakis]{lu2022FNOvsDeepONet}
Lu~Lu, Xuhui Meng, Shengze Cai, Zhiping Mao, Somdatta Goswami, Zhongqiang
  Zhang, and George~Em Karniadakis.
\newblock A comprehensive and fair comparison of two neural operators (with
  practical extensions) based on fair data.
\newblock \emph{Computer Methods in Applied Mechanics and Engineering},
  393:\penalty0 114778, 2022.

\bibitem[{Chen, Gengxiang} et~al.(2024){Chen, Gengxiang}, {Liu, Xu}, {Meng,
  Qinglu}, {Chen, Lu}, {Liu, Changqing}, and {Li,
  Yingguang}]{Chen2024riemannian}
{Chen, Gengxiang}, {Liu, Xu}, {Meng, Qinglu}, {Chen, Lu}, {Liu, Changqing}, and
  {Li, Yingguang}.
\newblock Learning neural operators on riemannian manifolds.
\newblock \emph{Natl Sci Open}, 3\penalty0 (6):\penalty0 20240001, 2024.
\newblock \doi{10.1360/nso/20240001}.
\newblock URL \url{https://doi.org/10.1360/nso/20240001}.

\bibitem[Li et~al.(2023)Li, Huang, Liu, and Anandkumar]{Li2023deform}
Zongyi Li, Daniel~Zhengyu Huang, Burigede Liu, and Anima Anandkumar.
\newblock Fourier neural operator with learned deformations for pdes on general
  geometries.
\newblock \emph{J. Mach. Learn. Res.}, 24\penalty0 (1), January 2023.
\newblock ISSN 1532-4435.

\bibitem[Lee et~al.(2021)Lee, Trask, Patel, Gulian, and Cyr]{lee2021POU_net}
Kookjin Lee, Nathaniel~A Trask, Ravi~G Patel, Mamikon~A Gulian, and Eric~C Cyr.
\newblock Partition of unity networks: deep hp-approximation.
\newblock \emph{arXiv preprint arXiv:2101.11256}, 2021.

\bibitem[Shazeer et~al.(2017)Shazeer, Mirhoseini, Maziarz, Davis, Le, Hinton,
  and Dean]{shazeer2017MoE_Sparse}
Noam Shazeer, Azalia Mirhoseini, Krzysztof Maziarz, Andy Davis, Quoc Le,
  Geoffrey Hinton, and Jeff Dean.
\newblock Outrageously large neural networks: The sparsely-gated
  mixture-of-experts layer.
\newblock \emph{arXiv preprint arXiv:1701.06538}, 2017.

\bibitem[Li et~al.(2008)Li, Perlman, Wan, Yang, Meneveau, Burns, Chen, Szalay,
  and Eyink]{li2008JHTDB}
Yi~Li, Eric Perlman, Minping Wan, Yunke Yang, Charles Meneveau, Randal Burns,
  Shiyi Chen, Alexander Szalay, and Gregory Eyink.
\newblock A public turbulence database cluster and applications to study
  lagrangian evolution of velocity increments in turbulence.
\newblock \emph{Journal of Turbulence}, \penalty0 (9):\penalty0 N31, 2008.

\bibitem[Graham et~al.(2016)Graham, Kanov, Yang, Lee, Malaya, Lalescu, Burns,
  Eyink, Szalay, Moser, et~al.]{graham2016JHTDB_channel}
J~Graham, K~Kanov, XIA Yang, M~Lee, N~Malaya, CC~Lalescu, R~Burns, G~Eyink,
  A~Szalay, RD~Moser, et~al.
\newblock A web services accessible database of turbulent channel flow and its
  use for testing a new integral wall model for les.
\newblock \emph{Journal of Turbulence}, 17\penalty0 (2):\penalty0 181--215,
  2016.

\bibitem[Perlman et~al.(2007)Perlman, Burns, Li, and
  Meneveau]{perlman2007JHTDB}
Eric Perlman, Randal Burns, Yi~Li, and Charles Meneveau.
\newblock Data exploration of turbulence simulations using a database cluster.
\newblock In \emph{Proceedings of the 2007 ACM/IEEE Conference on
  Supercomputing}, pages 1--11, 2007.

\bibitem[Blei et~al.(2017)Blei, Kucukelbir, and McAuliffe]{blei2017variational}
David~M Blei, Alp Kucukelbir, and Jon~D McAuliffe.
\newblock Variational inference: A review for statisticians.
\newblock \emph{Journal of the American statistical Association}, 112\penalty0
  (518):\penalty0 859--877, 2017.

\bibitem[Wang et~al.(2024)Wang, Li, Yuan, Peng, Liu, and
  Wang]{wang2024prediction}
Yunpeng Wang, Zhijie Li, Zelong Yuan, Wenhui Peng, Tianyuan Liu, and Jianchun
  Wang.
\newblock Prediction of turbulent channel flow using fourier neural
  operator-based machine-learning strategy.
\newblock \emph{Physical Review Fluids}, 9\penalty0 (8):\penalty0 084604, 2024.

\bibitem[Brown-Dymkoski et~al.(2014)Brown-Dymkoski, Kasimov, and
  Vasilyev]{brown2014characteristic}
Eric Brown-Dymkoski, Nurlybek Kasimov, and Oleg~V Vasilyev.
\newblock A characteristic based volume penalization method for general
  evolution problems applied to compressible viscous flows.
\newblock \emph{Journal of Computational Physics}, 262:\penalty0 344--357,
  2014.

\bibitem[Kadoch et~al.(2012)Kadoch, Kolomenskiy, Angot, and
  Schneider]{kadoch2012volume}
Benjamin Kadoch, Dmitry Kolomenskiy, Philippe Angot, and Kai Schneider.
\newblock A volume penalization method for incompressible flows and scalar
  advection--diffusion with moving obstacles.
\newblock \emph{Journal of Computational Physics}, 231\penalty0 (12):\penalty0
  4365--4383, 2012.

\bibitem[Engels et~al.(2015)Engels, Kolomenskiy, Schneider, and
  Sesterhenn]{engels2015numerical}
Thomas Engels, Dmitry Kolomenskiy, Kai Schneider, and J{\"o}rn Sesterhenn.
\newblock Numerical simulation of fluid--structure interaction with the volume
  penalization method.
\newblock \emph{Journal of Computational Physics}, 281:\penalty0 96--115, 2015.

\bibitem[Kolomenskiy and Schneider(2009)]{kolomenskiy2009fourier}
Dmitry Kolomenskiy and Kai Schneider.
\newblock A fourier spectral method for the navier--stokes equations with
  volume penalization for moving solid obstacles.
\newblock \emph{Journal of Computational Physics}, 228\penalty0 (16):\penalty0
  5687--5709, 2009.

\bibitem[Bertsekas(2014)]{bertsekas2014constrained}
Dimitri~P Bertsekas.
\newblock \emph{Constrained optimization and Lagrange multiplier methods}.
\newblock Academic press, 2014.

\bibitem[Kingma(2013)]{kingma2013VAEs}
Diederik~P Kingma.
\newblock Auto-encoding variational bayes.
\newblock \emph{arXiv preprint arXiv:1312.6114}, 2013.

\bibitem[Blundell et~al.(2015)Blundell, Cornebise, Kavukcuoglu, and
  Wierstra]{blundell2015weight}
Charles Blundell, Julien Cornebise, Koray Kavukcuoglu, and Daan Wierstra.
\newblock Weight uncertainty in neural network.
\newblock In \emph{International conference on machine learning}, pages
  1613--1622. PMLR, 2015.

\bibitem[Izmailov et~al.(2021)Izmailov, Vikram, Hoffman, and
  Wilson]{izmailov2021bayesian}
Pavel Izmailov, Sharad Vikram, Matthew~D Hoffman, and Andrew Gordon~Gordon
  Wilson.
\newblock What are bayesian neural network posteriors really like?
\newblock In \emph{International conference on machine learning}, pages
  4629--4640. PMLR, 2021.

\bibitem[Zhao et~al.(2023)Zhao, Gu, Varma, Luo, Huang, Xu, Wright, Shojanazeri,
  Ott, Shleifer, et~al.]{zhao2023FSDP}
Yanli Zhao, Andrew Gu, Rohan Varma, Liang Luo, Chien-Chin Huang, Min Xu, Less
  Wright, Hamid Shojanazeri, Myle Ott, Sam Shleifer, et~al.
\newblock Pytorch fsdp: experiences on scaling fully sharded data parallel.
\newblock \emph{arXiv preprint arXiv:2304.11277}, 2023.

\bibitem[Goyal(2017)]{imagenet1hour2017}
P~Goyal.
\newblock Accurate, large minibatch sg d: training imagenet in 1 hour.
\newblock \emph{arXiv preprint arXiv:1706.02677}, 2017.

\bibitem[Smith and Topin(2019)]{smith2019super}
Leslie~N Smith and Nicholay Topin.
\newblock Super-convergence: Very fast training of neural networks using large
  learning rates.
\newblock In \emph{Artificial intelligence and machine learning for
  multi-domain operations applications}, volume 11006, pages 369--386. SPIE,
  2019.

\bibitem[Bengio et~al.(2015)Bengio, Vinyals, Jaitly, and
  Shazeer]{bengio2015scheduled}
Samy Bengio, Oriol Vinyals, Navdeep Jaitly, and Noam Shazeer.
\newblock Scheduled sampling for sequence prediction with recurrent neural
  networks.
\newblock \emph{Advances in neural information processing systems}, 28, 2015.

\end{thebibliography}

\newpage
\appendix

\section{Appendix}

\subsection{Complex Values and Reparameterization Trick}\label{Complex_Reparameterization}
The complex-valued weights in the network architecture in Equation \eqref{eq:MOR_layer} are defined as
$ \theta_i = w_i+i\tilde w_i,$
where $\theta_i \in \mathbb{C}$, and the values $w_i \sim N(\mu_i, \sigma_i^2), \tilde w_i \sim N(\tilde \mu_i, \tilde \sigma_i^2)$, with Gaussian values parameterized by the reparameterization trick. This scheme yields twice as many parameters (1/2 real, and 1/2 imaginary) as compared to a comparable network parameterized with real-valued weights.

\section{Filtered DNS and LES fields from deterministic model}\label{sec:deteministic_les}

Although the main text describes a MFVI LES model, we include here a comparison between fields predicted by a deterministic model trained on the least squares loss and the fields from the filtered DNS. We note the close correspondence between the two sets of fields.

\begin{figure}[H]
    \centering
    \begin{minipage}[b]{1.0\linewidth}
        \centering
        \includegraphics[width=1\linewidth]{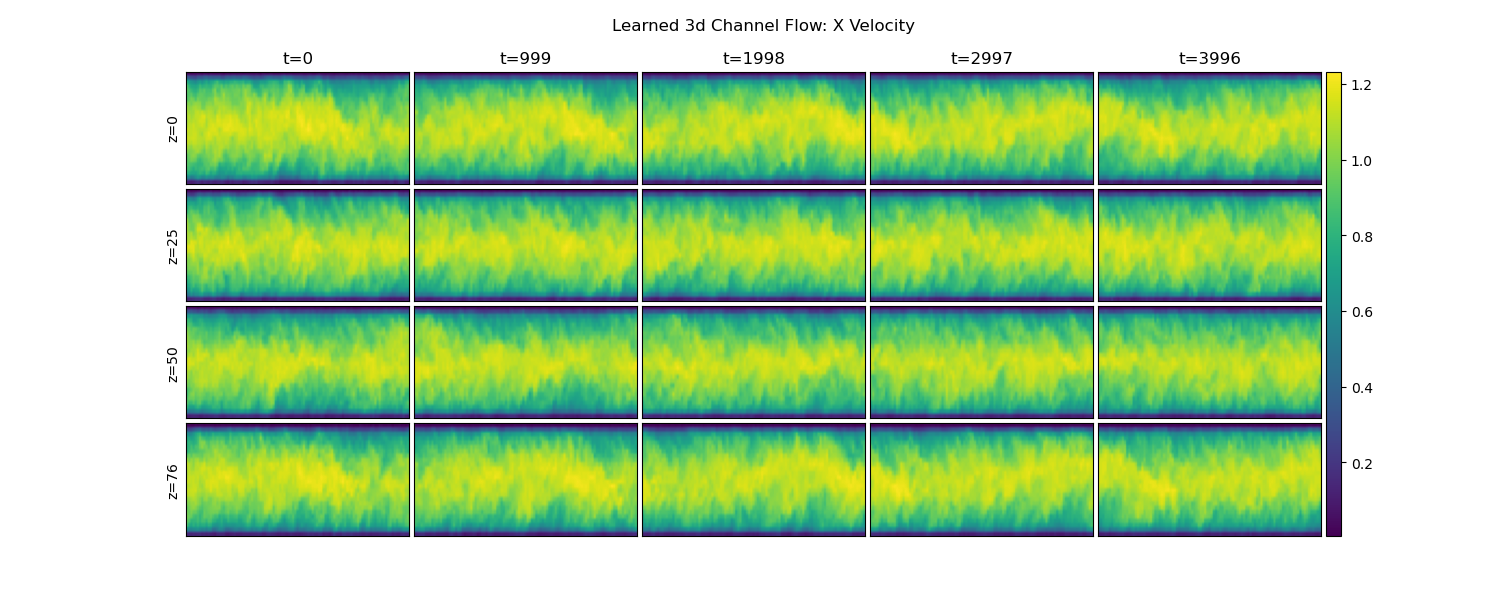}
        \caption{Learned Simulation: X Velocity}
    \end{minipage}
    \begin{minipage}[b]{1.0\linewidth}
        \centering
        \includegraphics[width=1\linewidth]{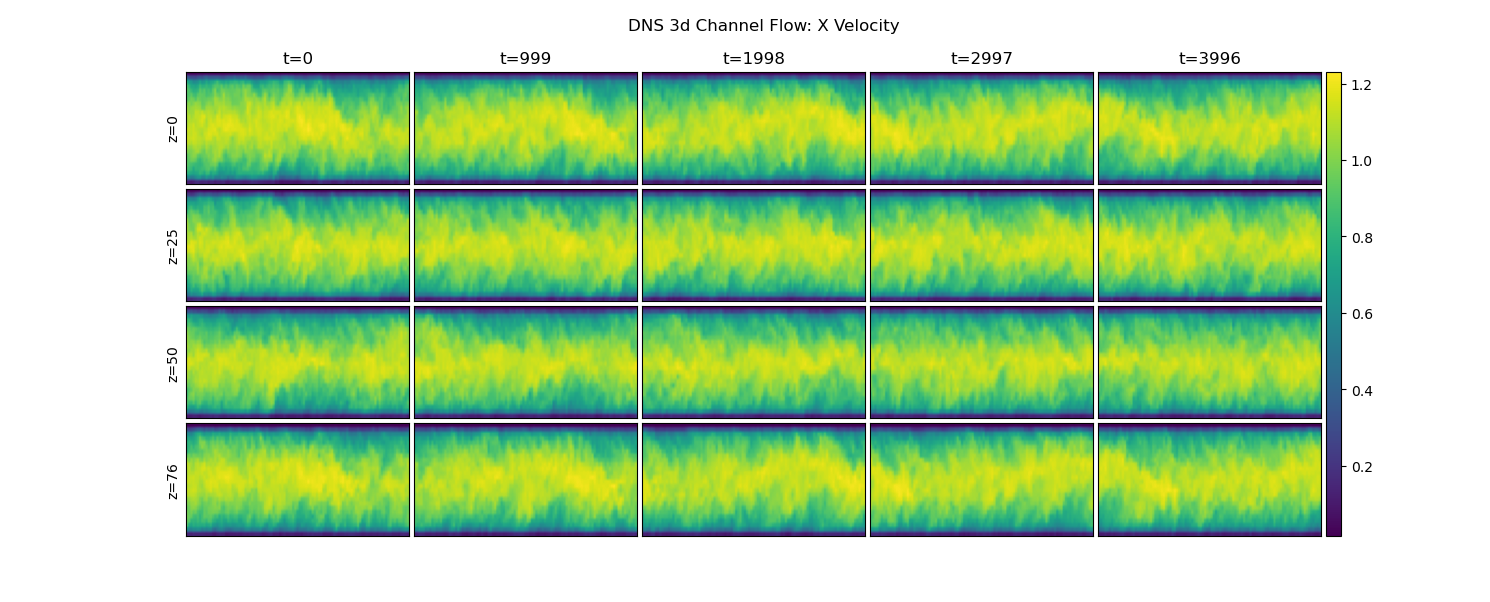}
        \caption{DNS: X Velocity}
    \end{minipage}
    \caption{Comparison of X velocity field from learned simulation vs DNS from JHTDB.}
    \label{fig:deterministic_x_velocity}
\end{figure}

\begin{figure}[H]
    \centering
    \begin{minipage}[b]{1.0\linewidth}
        \centering
        \includegraphics[width=1\linewidth]{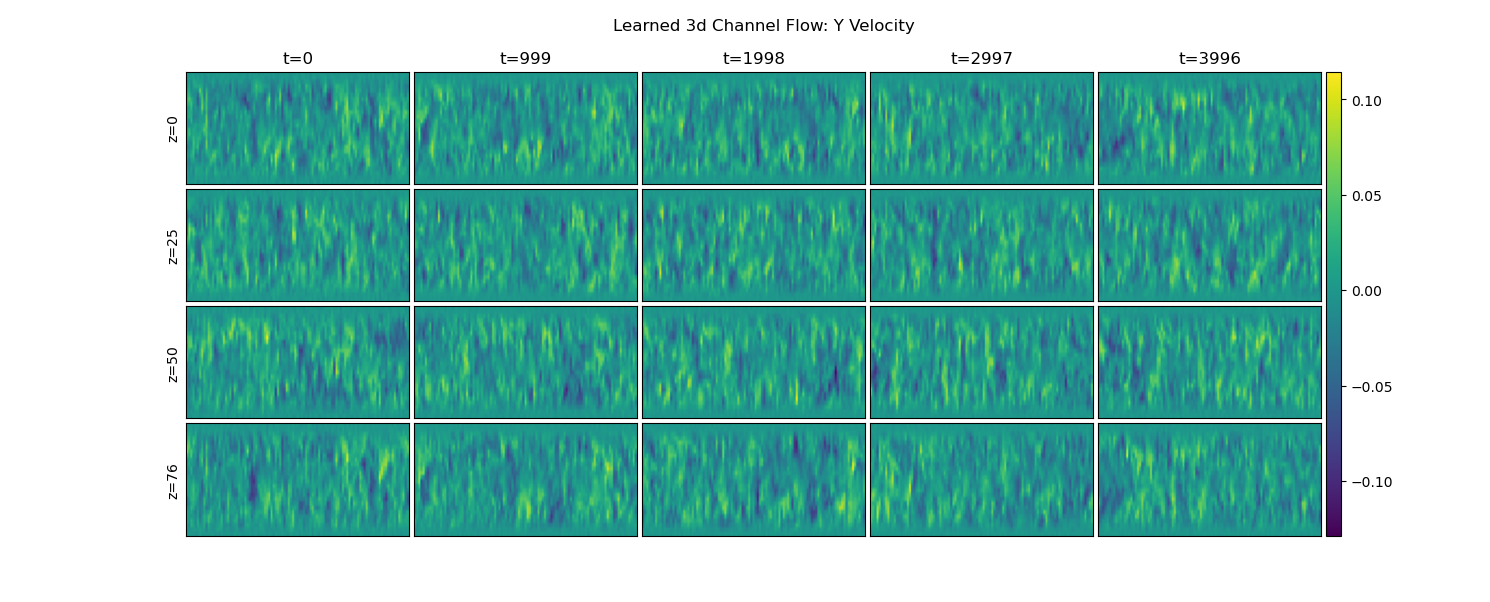}
        \caption{Learned Simulation: Y Velocity}
    \end{minipage}
    \begin{minipage}[b]{1.0\linewidth}
        \centering
        \includegraphics[width=1\linewidth]{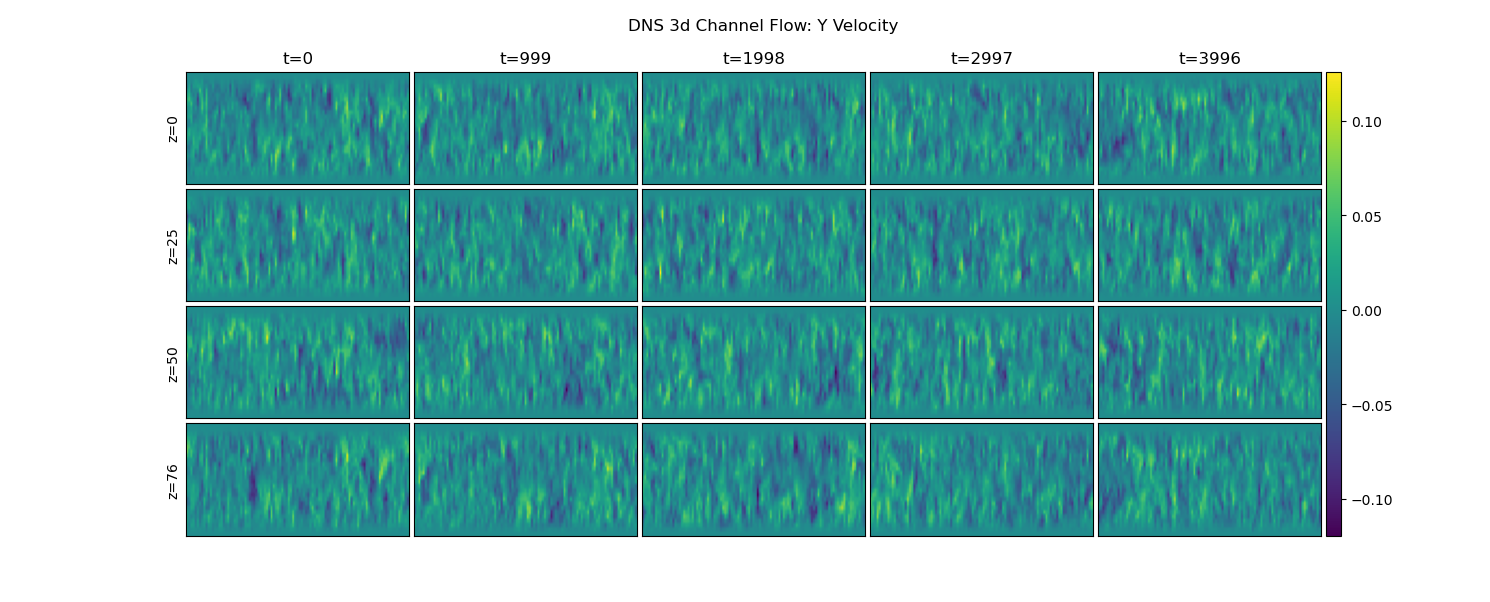}
        \caption{DNS: Y Velocity}
    \end{minipage}
    \caption{Comparison of Y velocity field from learned simulation vs DNS from JHTDB.}
    \label{fig:deterministic_y_velocity}
\end{figure}

\begin{figure}[H]
    \centering
    \begin{minipage}[b]{1.0\linewidth}
        \centering
        \includegraphics[width=1\linewidth]{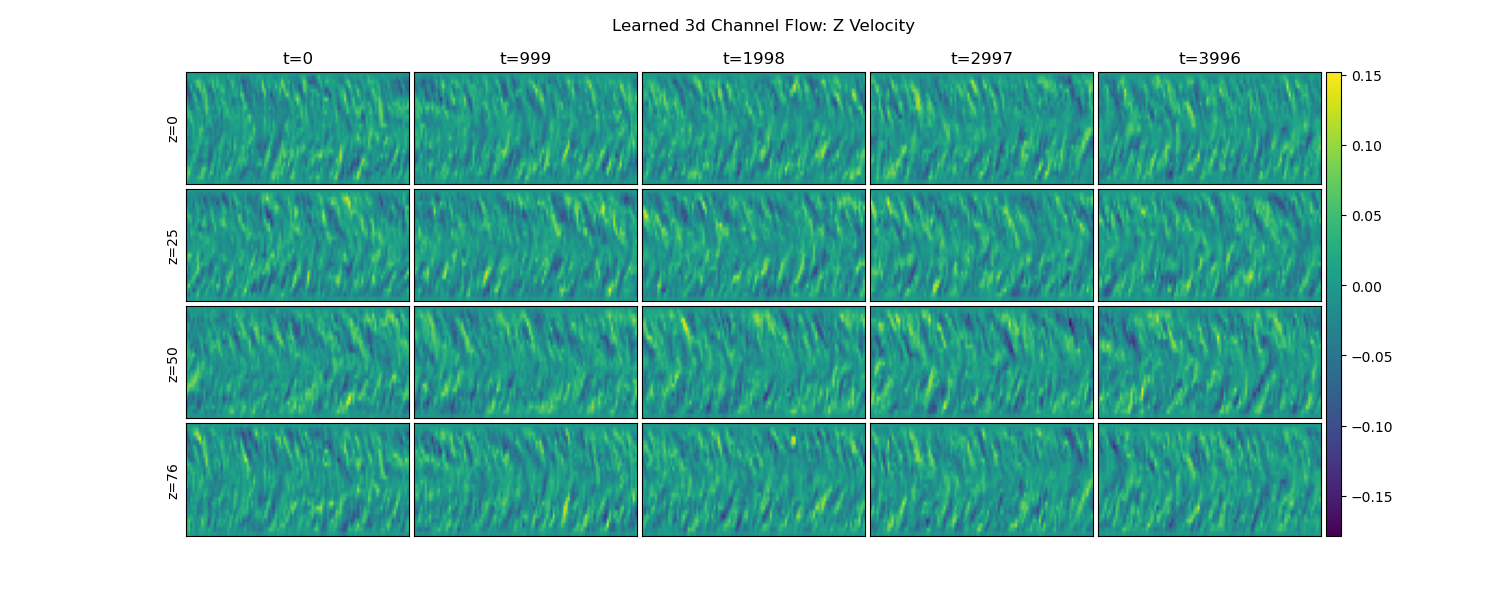}
        \caption{Learned Simulation: Y Velocity}
    \end{minipage}
    \begin{minipage}[b]{1.0\linewidth}
        \centering
        \includegraphics[width=1\linewidth]{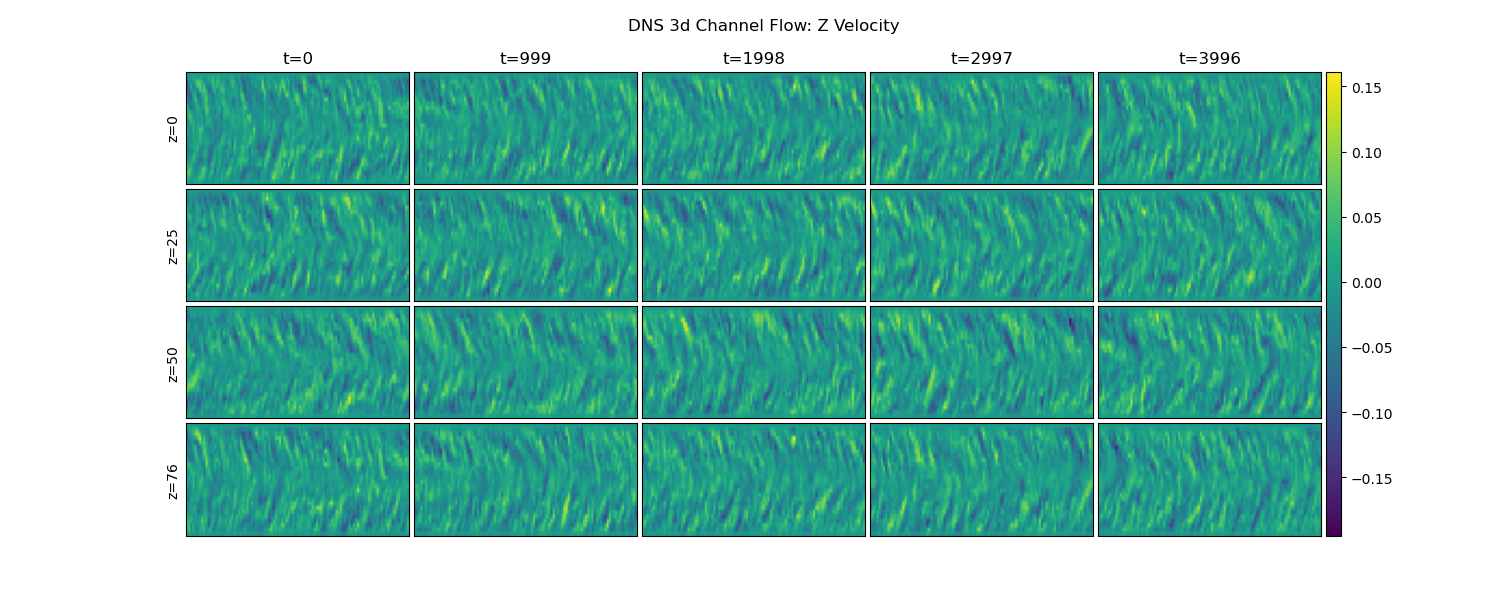}
        \caption{DNS: Y Velocity}
    \end{minipage}
    \caption{Comparison of Y velocity field from learned simulation vs DNS from JHTDB.}
    \label{fig:deterministic_z_velocity}
\end{figure}

\section{\textit{a priori} known physics for LES model}\label{sec:les_apriori}

Our LES model is an autoregressive model with \textit{a priori} known physics. See Section \ref{sec:autoregressive} for the general formulation of autoregressive models in our framework. In this section, we will start from the standard formulation of LES models and arrive at an LES model that incorporates a learned operator for various unclosed terms. As in Section \ref{sec:autoregressive}, we will separate the known physics from the unknown and expose the unknown physics to operator learning.

 Direct numerical simulation (DNS) solves the Navier-Stokes equations,
\begin{equation}\label{eq:ns}
\begin{aligned}
    &\partial_t v + \nabla {v} \otimes {u} = -\frac{1}{\rho} \nabla p + \nu \nabla^2 v & x \in \Omega\\
    &\nabla \cdot v = 0 & x \in \Omega \\
    &v = 0 & x \in \partial\Omega
\end{aligned}
\end{equation}
Equation \eqref{eq:ns} is typically expensive to integrate because real systems often contain a large range of spatiotemporal scales that must be resolved. The LES equations are obtained by applying a low pass spatiotemporal filter to the Navier-Stokes equations,
\begin{equation}\label{eq:les}
\begin{aligned}
    &\partial_t \tilde{v} + \nabla \tilde{v} \otimes \tilde{v} = -\frac{1}{\rho} \nabla \tilde{p} + \nu \nabla^2 \tilde{v} - \nabla \tau  & x \in \Omega \\
    &\nabla \cdot \tilde{v} = 0  & x \in \Omega \\
    &\tilde{v} = 0 & x \in \partial\Omega
\end{aligned}
\end{equation}
where $\tau = \widetilde{v \otimes v} - \tilde{v} \otimes \tilde{v}$ is the residual stress tensor. Since the small scale features have been filtered out, Equation \eqref{eq:les}, can be much cheaper to numerically integrate than Equation \eqref{eq:ns}. However, $\tau$ is an unclosed term that must be modeled. Traditionally, one uses a combination of intuition and analysis to arrive at a simple model with a few parameters that can be fitted to DNS simulation or experiments. Here, we will use POU-MOR-Physics to provide the closure.

To obtain our model, we will first consider LES in a triply periodic domain.
\begin{equation}\label{eq:les_periodic}
\begin{aligned}
    &\partial_t \tilde{v} + \nabla \tilde{v} \otimes \tilde{v} = -\frac{1}{\rho} \nabla \tilde{p} + \nu \nabla^2 \tilde{v} - \nabla \tau  & x \in \Omega \\
    &\nabla \cdot \tilde{v} = 0  & x \in \Omega 
    \end{aligned}
\end{equation}
We use the Chorin projection method to eliminate the pressure term and the explicit Euler time integrator to obtain an evolution operator,
\begin{equation}
\begin{aligned}
\hat{v}^{n+1} = &\tilde{v}^n  - \mathcal{F}^{-1} \left(i {\kappa} \cdot \mathcal{F}{\tilde{v}^n \otimes \tilde{v}^n} \right. \\
&\left. + \frac{i {\kappa}}{||{\kappa}||_2^2} ({\kappa} \otimes {\kappa}) : \mathcal{F}\tilde{v}^n \otimes \tilde{v}^n  - ||{\kappa}||_2^2 \mathcal{F}\tilde{{v}}^n \right)\\
= &\mathrm{LES}(\tilde{v}^n)
\end{aligned}
\end{equation}
where $\mathcal{F}$ is the Fourier transform and $\kappa$ is the wave vector.
 
This update operator does not include the boundary conditions, the missing sub-grid scale physics, or any forcing. We model these as an operator learned correction to the action above and obtain,
\begin{equation}
\tilde{v}^{n+1} = \mathcal{P} ( \mathrm{LES}(\tilde{v}^n)) = \mathcal{U}(\tilde{v}^n)
\end{equation}
where $\mathcal{P}$ is the POU-MOR-Physics operator discussed in Section~\ref{sec:model}. We can learn the closure $\mathcal{P}$ from low pass filtered DNS data and obtain a model that operates on much lower dimensional features than the original DNS, as we demonstrate in Section \ref{sec:les} and Appendix \ref{sec:deteministic_les}.

\end{document}